\pdfoutput=1

\documentclass[11pt]{article}

\usepackage[]{acl}

\usepackage{times}
\usepackage{latexsym}

\usepackage[T1]{fontenc}

\usepackage[utf8]{inputenc}

\usepackage{microtype}

\usepackage{amsfonts,amsmath,amssymb}
\usepackage{multirow}
\usepackage{graphicx}
\usepackage{adjustbox}
\usepackage{algpseudocode}

\title{Improving Downstream Task Performance by Treating Numbers as Entities}

\author{Dhanasekar Sundararaman$^1$, Vivek Subramanian$^2$, \\ \textbf{Guoyin Wang$^2$, Liyan Xu$^3$, Lawrence Carin$^1$} \\
  $^1$ Duke University \\
  $^2$ Amazon Alexa AI \\
  $^3$ Emory University \\
  \texttt{dhanasekar.sundararaman@duke.edu} \\}

\begin{document}
\maketitle
\begin{abstract}
Numbers are essential components of text, like any other word tokens, from which natural language processing (NLP) models are built and deployed. Though numbers are typically not accounted for distinctly in most NLP tasks, there is still an underlying amount of numeracy already exhibited by NLP models. In this work, we attempt to tap this potential of state-of-the-art NLP models and transfer their ability to boost performance in related tasks. Our proposed classification of numbers into entities helps NLP models perform well on several tasks, including a handcrafted Fill-In-The-Blank (FITB) task and on question answering using joint embeddings, outperforming the BERT and RoBERTa baseline classification.
\end{abstract}

\section{Introduction}

Named entity recognition (NER) is the task of classifying nouns within a document into categories to which they belong \cite{sang2003introduction}. Deep learning approaches for NER have been successful recently because of their ability to understand the underlying semantics of sentences \cite{shen2019learning} and incorporate this knowledge into tagging. Deep architectures for NER \cite{lample2016neural} include recurrent neural networks and attention based models including the Transformer \cite{vaswani2017attention}. 

\begin{figure}[t]
    \centering
    \includegraphics[width=\linewidth]{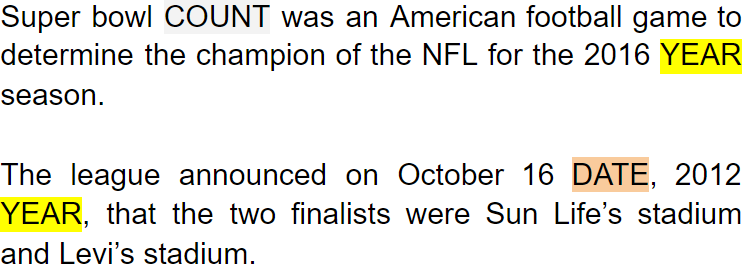}
    \caption{Sequence classification of numbers, or number entity recognition.}
    \label{fig:Nuer_intro}
\end{figure}

While NER has been performed on a number of datasets, tagging types of numbers, or \textit{number entities}, has not been a principal focus of NER. This is despite the fact that numbers are prevalent in most texts, comprising roughly 5\% of tokens in a sentence \cite{naik2019exploring}. Numbers can take on several categories, including years, ages, phone numbers, dates, etc. Figure \ref{fig:Nuer_intro} shows sample sentences in which numbers are classified into entity types. Distinguishing these different types of numbers can greatly enhance performance on downstream tasks \cite{sundararaman2020methods, sundararaman2022exploring}.

In this paper, we present several methods for addressing \textit{number entity recognition (NuER)}. We begin by demonstrating that state-of-the-art NLP models are able to classify types of numbers that appear in text \cite{wallace2019nlp, zhang2020language}. We study this by utilizing BERT \cite{devlin2018bert} to perform NuER on a custom version of the Stanford Question Answering Dataset (SQuAD) \cite{rajpurkar2016squad}, pruned to contain roughly 10K sentences for which the answers consist of numbers, and annotated in-house for numbers of six different entity types (year, percentage, date, count, age, and size) \cite{sundararaman2020methods}. We hereby refer to this as the SQuAD-Num dataset and the BERT classifier used to predict entities as the NuER model. We discover that BERT model performs well at this task on most of the categories listed above. The success of this model is made possible by the highly contextualized embeddings (\textit{e.g.,} the appearance of the name of a month, say March, next to a year, say 2003) inferred by BERT.

We then verify the effectiveness of BERT fine-tuned on SQuAD-Num to tag numbers that appear in a out-of-domain, human-annotated, 2K-sentence subset of the Numeracy 600K dataset \cite{chen2019numeracy}, which we call \textit{HA-Numeracy 2K}. We chose this dataset because of the abundance and diversity of numbers present. The subset was sampled in such a way as to maintain the distribution of number magnitudes that are originally present. We find that contextual embeddings are again beneficial in this setting, as surrounding tokens in both the source and target domains tend to be related. After verifying the quality of the annotations on the small sample, we annotated the full Numeracy 600K dataset using this model, which we hereafter call \textit{ENT-Numeracy 600K}. Table \ref{tab:data_percent} shows the number of samples in each entity type in SQuAD-Num. We share these annotations with the NLP community to further enhance research in this direction.

\begin{table}[t]
\centering
\begin{tabular}{|c|c|c|c|c|c|}
\hline
C & S & Y & P & D & A \\
\hline
\hline
1800 & 447 & 4355 & 291 & 418 & 82 \\
\hline
\end{tabular}
\vspace{-1ex}
\caption{Number of entities in our annotation for each type in SQuAD-Num. The table headers represent the following entity types respectively: Count, Size, Year, Percentage, Date, Age.}
\label{tab:data_percent}
\end{table}

Next, we jointly train embeddings for both tokens and number entities on the SQuAD-Num dataset. Answers to questions involving numbers are predicted by combining together token embeddings with trainable number entity embeddings. We find that this addition \cite{sundararaman2019syntax, sundararaman2021syntactic, subramanian2021lexical} enables BERT and RoBERTa models to better answer questions in SQuAD involving numbers. We later analyze the effectiveness of NuER through a handcrafted Fill-In-The-Blank (FITB) classification task. Our proposed NuER model helps the models classify numbers better when presented with the entity type information.

In summary, our contributions are as follows:
\begin{enumerate}
    \item We develop a BERT-based model for NuER which is used to tag numbers that appear in the SQuAD-Num dataset and in the out-of-domain HA-Numeracy 2K.
    \item We release datasets containing number entity annotations for the NLP community.
    \item We observe the effectiveness of NuER through a handcrafted Fill-In-The-Blank (FITB) task to predict a masked number in a sentence.
    \item We then jointly train embeddings for tokens and number entities, and show that numerical question answering performance improves when both are fed into the model.
\end{enumerate}

\section{Related Work}
Classification of numbers in plain text to entities has not been addressed methodologically and deserves considerable attention.  \cite{naik2019exploring} highlighted the importance of developing specialized embeddings for numbers such that basic properties including comparison and summation would be preserved. Since then In addition, \cite{sundararaman2020methods} designed a novel loss function that captured these properties and demonstrated improved performance on a variety of NLP tasks. Unlike their method which uses static embeddings, we propose to supplement the entity information of numbers in a contextual fashion to enable models to perform well on a number of tasks \cite{geva2020injecting}. There are existing methods on numerical word embeddings \cite{berg2020empirical, spithourakis2018numeracy}, but they didn't consider numbers as entities for the task of NER. NLP models have shown to use supplemental knowledge to its advantage \cite{tanwar2022unsupervised, sundararaman2019syntax, subramanian2021lexical}. Existing works have also shown that NLP models are intrinsically capable of numerical reasoning. This potential can be tapped using simple modifications like those proposed by \cite{andor2019giving}, in which the BERT model was used to perform arithmetic operations. Like the above-mentioned works, we seek to leverage the potential of existing NLP models to classify numbers into different entity types and transfer such ability across related tasks \cite{sundararaman2021learning} to boost performance.

\section{Preliminaries}
For each corpus, we index sentences by $n = 1, 2, \ldots, N$. Each token in a given vocabulary $\mathcal{V}$ is assigned a one-hot vector embedding $\mathbf{v} \in \left\{0, 1\right\}^{|\mathcal{V}|}$. Tokens in a sequence are indexed by $i = 1, 2, \ldots, T_n$, where $T_n$ is the length of sequence $n$ (after padding). During embedding lookup, tokens are mapped to embeddings $\mathbf{e}_{ni} \in \mathbb{R}^d$ ($d$ refers to the dimension of embedding). For all tasks, we use pre-trained models including BERT and RoBERTa to form latent representations of inputs; token embeddings are summed with position embeddings $\mathbf{p}_{ni} \in \mathbb{R}^{d}$ and segment embeddings $\mathbf{s}_{ni} \in \mathbb{R}^{d}$ prior to being fed into the Transformer encoder. The output consists of a classification token $\mathbf{c}_n \in \mathbb{R}^d$ and a sequence of contextualized embeddings $\mathbf{o}_{ni} \in \mathbb{R}^d$, one for each token. For each model, we use an Adam optimizer with $\eta = 0.001$, $\beta_1 = 0.9$, $\beta_2 = 0.999$, and $\epsilon = 10^{-7}$ to minimize the empirical loss across samples in each batch. More experimental settings are detailed in their respective sections. We train models on Tesla K80 GPUs and TPUs.

\section{Methods}
\subsection{Number entity recognition}
BERT, introduced by \cite{devlin2018bert}, has been shown to exhibit state-of-the-art performance for NER. Token embeddings are passed through a Transformer encoder, generating contextualized representations for each subword token, which are classified by the attention-based BERT classifier. We fine-tune the BERT model on the SQuAD-Num dataset. Both questions and context for each sequence are input to the NuER model, and the model outputs a sequence of tags denoting whether each token is one of the six types of numbers or ``other'' (seven total classes). The outputs of the NuER model for each element in the sequence, given by the $\mathbf{o}_i$, are mapped through a linear layer (we drop the $n$ subscript for clarity), and the resultant scalars are fed into a softmax function:
\begin{align}
    \mathbf{\hat{y}}_i = \frac{\exp(\mathbf{W}\mathbf{o}_i)}{\sum_{k=1}^K \exp(\mathbf{w}_k^\top\mathbf{o}_i)}
\end{align}
where $\mathbf{w} \in \mathbb{R}^{d}$, $\mathbf{W} = \left[\mathbf{w}_1, \ldots, \mathbf{w}_k, \ldots, \mathbf{w}_K\right]^\top \in \mathbb{R}^{K\times d}$ are the parameters of the linear layer, $k = 1, \ldots, K$ indexes the entity labels, and the $\exp$ operator is applied elementwise. The model is trained with categorical cross-entropy loss:
\begin{align}
\mathcal{L} = \sum_{i=1}^T\sum_{k=1}^K -y_{ik}\log(\hat{y}_{ik})
\end{align}
where $y_{ik} \in \left\{0, 1\right\}$ denotes whether the $i$th element belongs to class $k$. We divide SQuAD-Num into 75\% training, 10\% validation, and 15\% testing. We fine-tune the NuER model for 20 epochs on SQuAD-Num and evaluate our model on the out-of-domain HA-Numeracy 2K.

\subsection{Jointly trained embeddings}
We experiment with training embeddings for tokens jointly with number entities, referred as Jointly trained Embeddings (JEM) \cite{wang2018joint}. Specifically, we compare the performance of numerical question answering on SQuAD-Num, with and without entity embeddings. As input to the pretrained model, we feed in pairs $(\mathbf{v}_i, \mathbf{z}_i)$ of tokens $\mathbf{v}_i$ and number entity labels $\mathbf{z}_i$. $\mathbf{z}_i \in \left\{0, 1\right\}^{K+1}$ is a one-hot vector, where $K=6$ indicates the six entities of numbers in the corpus. During embedding lookup, the pairs of embeddings for each token are summed and fed as inputs:
\begin{align}
    \mathbf{\tilde{e}}_i = \mathbf{e}_i + \mathbf{h}_i
\end{align}
where $\mathbf{h}_i$ denotes the trainable number entity embedding for token $i$. We compare the performance of this model after fine-tuning for two epochs against the baseline, which is only fed the token identities as input. As is standard for SQuAD, the model is trained with cross-entropy loss in order to predict the probability of any token in the context being the start/end of a span that answers the numerical question:
\begin{align}
\mathcal{L} = \sum_{i=1}^T -y_{i}\log(\hat{y}_{i})
\end{align}
where here, $y_i \in \{0,1\}$ identifies the start/end of a span and $\hat{y}_i$ is the corresponding probability predicted by our model.

\begin{table*}[t]
\centering
\begin{tabular}{|c|c||c|c|c|c|c|c|c|}
\hline
                               &           & Year  & Count & Percentage & Age   & Size  & Date   & Total \\ \hline\hline
\multirow{3}{*}{SQuAD-Num}         & Precision & 98.61 & 88.66 & 95.08      & 94.12 & 87.84 & 100.00 & 95.38 \\ \cline{2-9} 
                               & Recall    & 98.99 & 95.91 & 96.67      & 88.89 & 84.42 & 100.00 & 97.24 \\ \cline{2-9} 
                               & F1        & 98.8  & 92.15 & 95.87      & 91.43 & 86.09 & 100.00 & 96.3  \\ \hline\hline
\multirow{3}{*}{ENT-Numeracy 600K} & Precision & 97.96 & 92.00 & 91.67      & 75.00 & 54.55 & 96.88  & 94.26 \\ \cline{2-9} 
                               & Recall    & 96.48 & 81.01 & 95.65      & 7.14  & 30.00 & 85.94  & 84.14 \\ \cline{2-9} 
                               & F1        & 97.21 & 86.16 & 93.62      & 13.04 & 38.71 & 91.09  & 88.91 \\ \hline
\end{tabular}
\caption{Performance metrics for NuER on SQuAD-Num and ENT-Numeracy 600K for each of the six entity types.}
\label{tab:numentreg}
\end{table*}

\subsection{FITB}

In this subsection, we measure the effectiveness of NuER on classification tasks. We handcraft a Fill-In-The-Blank (FITB) task of predicting a masked number in a given input sentence.  This is challenging as there are several thousand unique numbers in the corpus to which a masked number could be classified. It measures the true capability of BERT to fill in a number given the context. Our model which we refer to as $BERT_{NuER}$ consists of BERT tokenizer with six special tokens namely \textit{<size>, <age>, <count>, <date>, <year>,} and \textit{<percentage>}. These special tokens are added next to the CLS as an indication to the model about their respective number entity types. On the classification side, a linear projection layer of size $M \times D$ is created to learn about the six different entity types ($M=6$, $D=768$). These projections are then concatenated with the BERT hidden states of 768 dimensions to obtain final set of dimensions $(768*2) \times M$ which is projected again to a classification layer of $V$ output vocabulary items. Here, the $V$ output vocabulary items represent the unique numbers that are found in the corpus.

\begin{table}[t]
\centering
\begin{tabular}{|c|c||c|c|}
\hline
                               &           & Year  & Count  \\ \hline\hline
\multirow{3}{*}{Few-shot}         & Precision & 87.09 & 76.27 \\ \cline{2-4} 
                               & Recall    & 97.32 & 89.59 \\ \cline{2-4} 
                               & F1        & 91.92 & 82.39  \\ \hline\hline
\multirow{3}{*}{Full finetuning} & Precision & 91.15 & 80.53 \\ \cline{2-4} 
                               & Recall    & 98.49 & 95.33  \\ \cline{2-4} 
                               & F1        & 94.68 & 87.31  \\ \hline
\end{tabular}
\caption{Few-shot transfer vs fine-tuning performance on NuER entity types.}
\label{tab:fewshot}
\end{table}

\section{Results and Discussion}

\subsection{Number entity recognition}
Table \ref{tab:numentreg} shows the performance of our NuER model on each dataset. We find that the model performs well across all categories on SQuAD-Num and in most categories of ENT-Numeracy 600K. Performance drops are observed for the ``Age'' and ``Size'' categories, partially because there are not enough training instances of these two categories in SQuAD-Num. For ``Age,'' precision is high while recall is low. This is a limitation of our work and could be because the numbers that are classified as ``Age'' can typically be confused with other entities including ``Size,'' suggesting that the model is only able to capture age accurately when it is quite obvious in the dataset. For ``Size,'' both precision and recall are low,  perhaps because, unlike the other types of numbers, size may have many different units (\textit{e.g.,} feet, meters, acres, \textit{etc.}), which make it more difficult to associate with the numerical value.  Table \ref{tab:fewshot} shows that the classification performance when only few of the samples are used in fine-tuning is still comparable to the full fine-tuning performance on two classes. Though it works well for few classes, the F1 scores drop significantly for classes like Age and Size.

\subsubsection{Training settings}
For Table \ref{tab:numentreg} experiments, the learning rate was set to $2\times 10^{-5}$. For few-shot used in Table \ref{tab:fewshot}, only first 20 samples (using grid search, fewer samples to retain maximum accuracy compared to full finetuning) from Year and Count, while full fine-tuning uses all the samples.

\begin{table}[t]
\begin{center}
\begin{tabular}{|c|c|c|c|}
    \hline
    Model & Version & Exact match & F1 \\
    \hline
    \multirow{2}{*}{$BERT_{B}$}& Baseline & 90.66 & 90.71\\
    &JEM & \textbf{91.17} & \textbf{91.33} \\
    \hline
    \multirow{2}{*}{$RoBERTa_{B}$}& Baseline & 95.83 & 95.88\\
    &JEM & \textbf{96.84} & \textbf{96.95} \\
    \hline
    \multirow{2}{*}{$BERT_{L}$}& Baseline & 93.19 & 93.28\\
    &JEM & \textbf{94.83} & \textbf{95.01} \\
    \hline
\end{tabular}
\caption{Improvements on SQuAD-Num dataset using Jointly trained Embedding (JEM) technique with $BERT_{B}$ (base), $BERT_{L}$ (large), and $RoBERTa_{B}$ (base). All numbers are tested for significance and have a standard deviation of 0.05.} 
\label{tab:jointtrain}
\end{center}
\end{table}

\begin{figure*}[t]
    \centering
    \includegraphics[width=\linewidth]{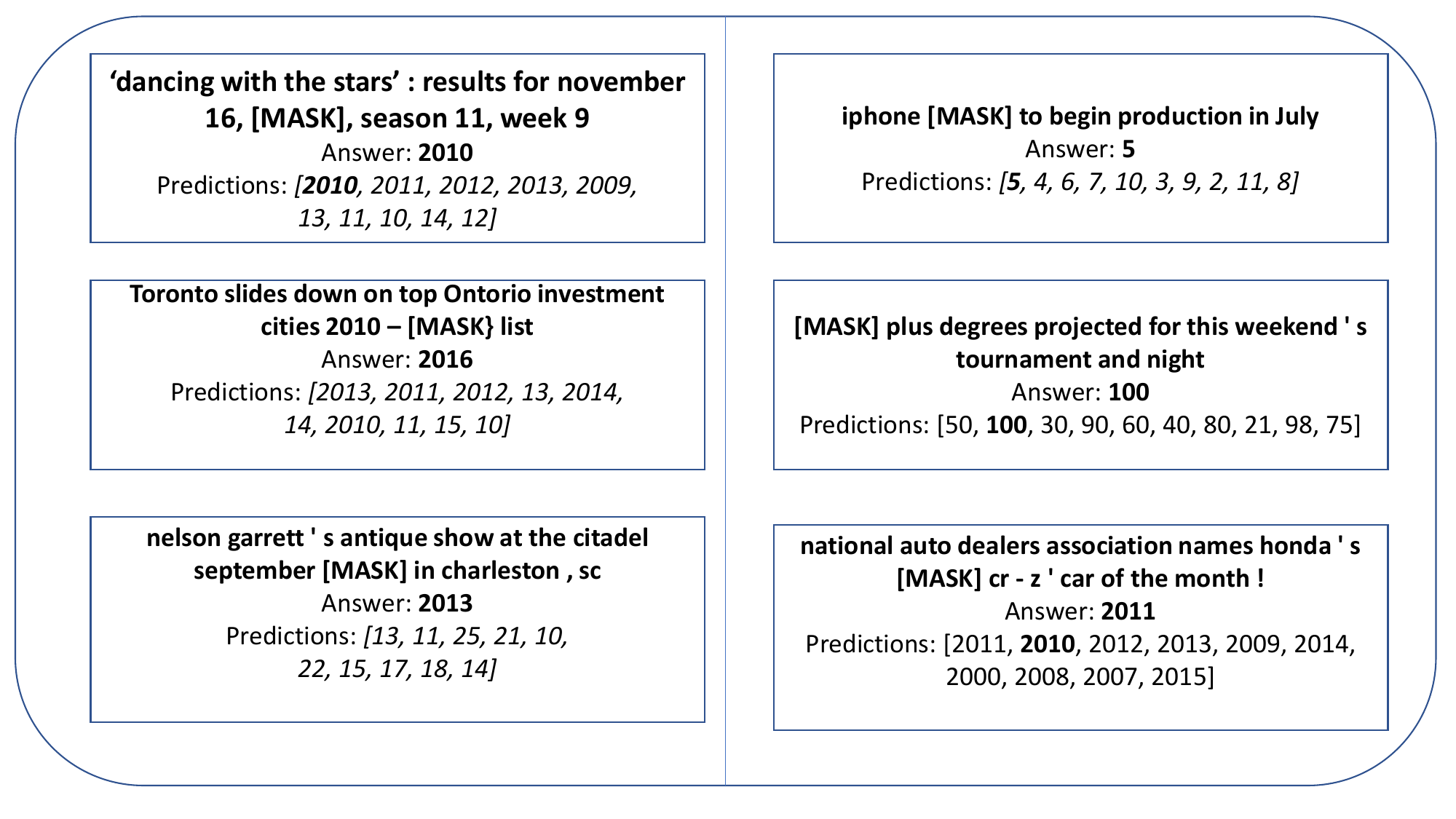}
    \caption{Qualitative comparison of predictions of numbers on the FITB task between the BERT base and $BERT_{NuER}$.  The left column shows BERT base predictions, while the right columns shows $BERT_{NuER}$ predictions. $BERT_{NuER}$ predictions align with a number entity more than BERT base.}
    \label{fig:qual_comp}
\end{figure*}

\subsection{Jointly trained embeddings}

Table \ref{tab:jointtrain} compares the performance of our model trained on numerical question answering against the BERT baseline. Our model demonstrates gains in both the exact match and F1 scores, suggesting that the additional information imparted by knowledge of the type of numbers that are present in the question and context aids the model in localizing the number within the context. A variety of models including $BERT_B$ (base), $BERT_L$ (large) and $RoBERTa_B$ (base) \cite{devlin2018bert,liu2019roberta} are used to understand if these improvements are consistent. We found that in terms of overall performance, $RoBERTa_{B} > BERT_{L} > BERT_{B}$, with $RoBERTa_B$ being the highest, while in terms of performance improvements, $BERT_{L} > RoBERTa_{B} > BERT_{B}$ with $BERT_L$ JEM obtaining an improvement of 1.64 exact match and 1.7 F1 points respectively.

\subsubsection{Training settings}
BERT for QA consists of a pretrained model along with classification heads for start and ends of spans. The model was trained for 2 epochs with a learning rate of $2\times 10^{-5}$. During training, embeddings for numbers and their entities are learned in parallel.

\subsection{FITB}
Table \ref{tab:FITB} shows the performance improvements of $BERT_{NuER}$ over the BERT base for FITB. The top-$k$ performance tells us that $BERT_{NuER}$ is able to predict the correct answer within the top-$k$ entries. We see that the gap in performance improvement increases as $k$ increases. This tells us that the enriched model is more accurate in its predictions. The dist metric quantifies the average absolute distance between pairs of actual value and the predicted values. This ensures the quality of numeral predictions. For example, as shown in the first column of Figure \ref{fig:qual_comp}, when the ground truth is a Year, we expect all the other predictions to also be years (as opposed to Count or other classes). The dist metric calculates the spread of numbers while top-$k$ calculates the accuracy itself. We see that the dist is significantly lower than the baseline model. Figure \ref{fig:qual_comp} shows that $BERT_{NuER}$ generates top-$k$ predictions whose elements are more consistent with the desired class label.

\subsubsection{Training settings}
For the FITB experiment, DistilBERT \cite{sanh2019distilbert} with an Adam optimizer and a learning rate of $2\times 10^{-4}$ is used. The model is trained for 3 epochs with a batch size of 64. On the architectural side, adding a linear layer to the BERT hidden states yielded marginal improvements. But, concatenating the linear layer responsible for learning number entity onto the BERT hidden states followed by classification lead to significant improvements.

\begin{table}[t]
\begin{adjustbox}{width=\columnwidth}
\centering
\begin{tabular}{|c|c|c|c|c|}
\hline
            & Top-1 & Top-2 & Top-5 & Top-10     \\ \hline
Baseline     & 36.88    & 52.51 & 71.57 & 79.84 \\ \hline
Dist     & 624    & 1279.06 & 3275.22 & 8240.56 \\ \hline
Ours & \textbf{37.69} & \textbf{53.9} & \textbf{73.1} & \textbf{81.47} \\ \hline
Dist     & \textbf{554}    & \textbf{1110} & \textbf{2829.73} & \textbf{6435.14} \\ \hline
\end{tabular}
\end{adjustbox}
\caption{Predictive performance on the FITB task. All numbers are tested for significance and have a standard deviation of 0.01}
\label{tab:FITB}
\end{table}

\section{Conclusions}
We have defined number entities and find that this aids multiple NLP tasks, via joint training, and classification. We also introduce a dataset representing number entities and a methodology to annotate unseen numbers. In the classification objective, we find that the addition of entities provides a considerable boost in FITB evident through the quality of predictions. Through our datasets and methods, we hope to incite further research at the intersection of numeracy and NER.

\section*{Acknowledgments}
The authors would like to thank Vishwanath Seshagiri for helping with the annotations on the SQuAD-Num dataset.

\bibliography{anthology,custom}
\bibliographystyle{acl_natbib}

\end{document}